\documentclass[10pt,twocolumn,letterpaper]{article}

\usepackage[pagenumbers]{cvpr}

\usepackage{cuted}

\usepackage{currfile}

\usepackage{caption}

\definecolor{cvprblue}{rgb}{0.21,0.49,0.74}
\usepackage[pagebackref,breaklinks,colorlinks,allcolors=cvprblue]{hyperref}

\def\confName{CVPR}
\def\confYear{2027}

\usepackage[T1]{fontenc}

\usepackage{amsmath,amssymb,amsthm}

\usepackage{graphicx}
\usepackage{xcolor}
\usepackage{pgfplots}
\pgfplotsset{compat=1.18}
\usepackage{stfloats}

\usepackage{booktabs,multirow,tabularx,array}

\usepackage{tikz}

\providecolor{AblDefault}{RGB}{233,229,244}

\providecommand{\pmmMemory}{\mathcal{M}}

\usepackage{colortbl}
\title{Programmatic Multimodal Memory: Learning to Read Long Videos with Code}

\author{
Xuanming Shang\textsuperscript{1,2,*} \quad
Yuhang Han\textsuperscript{1,2,3,*} \quad
Yan Shu\textsuperscript{2,4,*} \quad
Dingning Liu\textsuperscript{2,5} \quad
Li Ding\textsuperscript{1,2}\\
Tao Qiang\textsuperscript{1,2} \quad
Zhiqi Shao\textsuperscript{6} \quad
Zequn Yang\textsuperscript{1,2,\ensuremath{\ddagger}} \quad
Yanfeng Wang\textsuperscript{1,\ensuremath{\dagger}}\\[3pt]
{\small \textsuperscript{1}Shanghai Jiao Tong University \quad
\textsuperscript{2}Shadoweave \quad
\textsuperscript{3}The Hong Kong University of Science and Technology (Guangzhou)}\\
{\small \textsuperscript{4}University of Trento \quad
\textsuperscript{5}University of Electronic Science and Technology of China \quad
\textsuperscript{6}Chongqing University}\\[2pt]
{\small \texttt{\{sxm2021, wangyanfeng622\}@sjtu.edu.cn}}\\[-4pt]
}

\begin{document}
\maketitle
{\renewcommand{\thefootnote}{}\footnotetext{\textsuperscript{*}Equal contribution. \textsuperscript{\ensuremath{\dagger}}Corresponding author. \textsuperscript{\ensuremath{\ddagger}}Project lead.}}
\begin{abstract}
Answering questions about hours- or days-long videos requires finding a few seconds of evidence in recordings far longer than any model can read at once. Most memory systems decide what matters at write time. Before any question arrives, they consolidate the video into compressed memory banks or structured graphs and lose details that later questions need. We instead defer this decision to read time. Programmatic Multimodal Memory (PMM) writes a multi-scale temporal memory of timestamped captions and transcripts once, without consolidation, and keeps it aligned with the raw video. Single-round retrieval misses much of the evidence, so a Root model reads this memory with code over multiple turns in a persistent Python REPL. Its programs filter records across days, query a frozen submodel in batches, and inspect raw frames on demand. The Root now decides what matters, yet small open-weight models struggle zero-shot to coordinate code search and frame inspection. We therefore train Qwen3.5-9B as the Root in this REPL with evidence-grounded supervised fine-tuning and outcome-reward reinforcement learning. The resulting PMM-9B gains 12.1 points over its zero-shot version and outperforms every open-source model and agent baseline on all five personal-memory and long-video benchmarks. It also beats standalone GPT-5 and Gemini 2.5 Pro on average, and by nearly 20 points on week-long EgoLifeQA. Open-source baselines that use more input tokens per question than PMM-9B are all less accurate, and those that use fewer score at least 16.6 points lower. Project page: \url{https://shadow-weave.github.io/PMM/}.
\end{abstract}

\section{Introduction}
\label{sec:intro}
\begin{figure}[t]
\centering
\includegraphics[width=\linewidth]{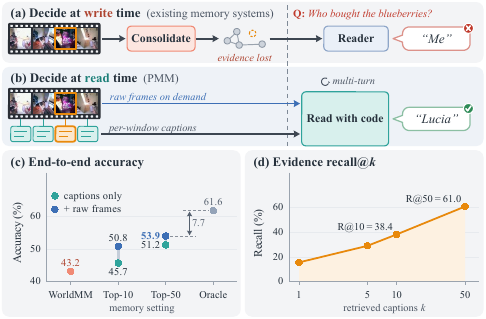}
\caption{\textbf{Deciding what matters at write time vs.\ read time.}
\textbf{(a, b)} Consolidating at write time can lose evidence that a later question needs, whereas PMM captions every window without merging and reads this temporal memory with code at read time, inspecting raw frames on demand.
\textbf{(c, d)} On 484 EgoLifeQA questions with annotated evidence (Qwen3-VL-8B reader), single-round caption retrieval matches WorldMM~\cite{yeo2026worldmm} and gains from raw frames, but stays below the Oracle (c) because evidence recall is low (d).}
\label{fig:teaser}
\vspace{-10pt}
\end{figure}

With wearable cameras, assistants can remember what their users saw and answer questions such as who bought the blueberries~\cite{yang2025egolife,tian2026ego}. Answering such questions requires finding a few seconds of evidence in recordings that span hours or days, far longer than any Video LLM can read at once. Even models with million-token contexts~\cite{team2024gemini,shu2025video} cannot hold a week of video, and agents that search raw frames from scratch for every question~\cite{wang2024videoagent,ye2025re,rauf2026hpp} become too costly as recordings grow. Long-video understanding therefore depends on persistent memory. Most memory systems decide what matters at write time (Fig.~\ref{fig:teaser}a). Before any question arrives, they consolidate the video into compressed memory banks~\cite{song2024moviechat,he2024ma} or structured memories and graphs~\cite{yang2025egolife,yeo2026worldmm,long2026seeing}. Some also add visual lookups~\cite{yeo2026worldmm,tian2026ego}.

Consolidating video at write time loses details that later questions need~\cite{yeo2026worldmm,rauf2026hpp}, because the writer cannot know whether a future question will ask about a week-long habit or a fleeting visual cue. We instead defer this decision to read time (Fig.~\ref{fig:teaser}b), keeping an unconsolidated temporal memory aligned with the raw video. In a controlled study on EgoLifeQA~\cite{yang2025egolife}, single-round retrieval over 30-second captions already matches the consolidated memory of WorldMM~\cite{yeo2026worldmm}, and inspecting raw frames of the retrieved segments adds 5.1 points (Fig.~\ref{fig:teaser}c). Yet unconsolidated memory leaves clues scattered across hours or days. Single-round retrieval recalls under 40\% of the evidence in the top 10, and widening it to the top 50 multiplies visual cost fivefold while still missing nearly 40\% (Fig.~\ref{fig:teaser}d). Deciding at read time thus requires adaptive search. However, recent tool-use agents~\cite{yeo2026worldmm,tian2026ego} rely on fixed query interfaces and turn-by-turn calls that accumulate retrieved text in the prompt and cannot compose many lookups in one turn.

To decide what matters at read time, we introduce Programmatic Multimodal Memory (PMM), which pairs a multi-scale temporal memory with programmatic reading in a persistent Python REPL. At write time, PMM builds this question-agnostic memory once without consolidation, storing windowed, timestamped captions and transcripts aligned with the raw video. At read time, PMM exposes the memory as a REPL variable rather than prompt text, building on Recursive Language Models (RLMs)~\cite{zhang2025recursive}. Instead of calling fixed query interfaces, a Root model reads it with code over multiple turns, conditioning each turn on what earlier programs printed. Its programs filter records across days and scales, query a frozen submodel in batches on selected text (\texttt{ask}), and use the same submodel to inspect raw frames on demand (\texttt{see}). By narrowing candidate intervals in text before calling \texttt{see}, the Root inspects frames only where needed. A single program can also loop over many candidate moments and aggregate submodel outputs in code, so one turn evaluates many clips while the Root context stays compact, even for week-long videos.

The Root now decides what matters, yet small open-weight models struggle zero-shot to coordinate code search and frame inspection. We therefore train Qwen3.5-9B~\cite{qwen2026qwen35} as the Root in the same REPL used at inference. Supervised fine-tuning uses only correct teacher trajectories that observed the annotated evidence, and reinforcement learning with GRPO~\cite{shao2024deepseekmath} then uses answer correctness as the only reward. Across EgoLifeQA, HippoVlog~\cite{lin2026hippomm}, MLVU~\cite{zhou2025mlvu}, LongVideoBench~\cite{wu2024longvideobench}, and VideoMME Long~\cite{fu2025video}, the resulting PMM-9B averages 72.70\%, gaining 12.1 points over its zero-shot version. PMM-9B outperforms every open-source model and agent baseline on all five benchmarks. It also beats standalone GPT-5~\cite{openai2025gpt5} and Gemini 2.5 Pro~\cite{comanici2025gemini} on average, and by nearly 20 points on week-long EgoLifeQA. At read time, PMM-9B uses 158K input tokens per question across all model calls (Fig.~\ref{fig:efficiency}). Open-source baselines that use more tokens are all less accurate, and those that use fewer score at least 16.6 points lower. Replacing the Root with DeepSeek-V4.1-Flash~\cite{deepseek2026v41} raises PMM to 75.82\%, the highest of all systems.

In summary, we make three contributions.
\begin{itemize}
    \item We propose PMM, which defers the decision of what matters to read time. PMM writes a temporal memory once without consolidation, and the Root reads it with code over multiple turns and inspects raw frames on demand.
    \item We train Qwen3.5-9B as the Root in the inference REPL with evidence-grounded supervised fine-tuning and outcome-reward reinforcement learning. The resulting PMM-9B gains 12.1 points over its zero-shot version.
    \item On all five personal-memory and long-video benchmarks, PMM-9B outperforms every open-source model and agent baseline and beats standalone GPT-5 and Gemini 2.5 Pro on average, while open-source baselines using fewer input tokens score at least 16.6 points lower.
\end{itemize}

\section{Related Work}
\label{sec:related}

\noindent\textbf{Long video understanding and memory.}
Existing Video LLMs scale context lengths~\cite{team2024gemini,zhang2024long,chen2025longvila,shu2025video} or search frames and clips with agents~\cite{wang2024videoagent,wang2025videotree,ye2025re,eltahir2026videoatlas}, some guided by reinforcement learning~\cite{pan2026timesearch,tian2026ego}. HPP~\cite{rauf2026hpp} lets a coding LLM probe the video by calling a small VLM, but it searches raw frames from scratch for every question without a persistent memory. As benchmarks grow from minutes~\cite{mangalam2023egoschema} and hours~\cite{fu2025video,wang2025lvbench,chandrasegaran2024hourvideo} to week-long recordings~\cite{yang2025egolife,tian2026ego}, videos exceed any context window and per-question search becomes too costly, so systems build memory as the video arrives. Memory-augmented models compress visual tokens into fixed-size memory banks~\cite{song2024moviechat,he2024ma,zhang2025flash,chen2024videollm}, whereas retrieval-based systems convert the video into textual captions, knowledge graphs, event logs, or entity-centric stores~\cite{ren2026videorag,yang2025egolife,lin2026hippomm,long2026seeing,yeo2026worldmm,sun2026egograph,chen2026em2mem}. These systems decide what matters at write time and consolidate the video before any question arrives, so details dropped at write time cannot be recovered from the memory at read time. WorldMM~\cite{yeo2026worldmm} and Ego-R1~\cite{tian2026ego} add visual lookups, but they use fixed query interfaces and one tool call per turn. PMM instead keeps an unconsolidated temporal memory aligned with the raw video, and the Root decides what matters at read time. A single program locates candidate moments across days in the temporal memory and composes many \texttt{see} calls on the raw frames of these moments.

\noindent\textbf{Programmatic and recursive reasoning.}
Visual programming composes pretrained perception modules into executable programs~\cite{gupta2023visual,suris2023vipergpt}, and later work extends this idea to video question answering on short clips by prompting a frozen LLM~\cite{choudhury2024video,min2024morevqa}. CodeAct~\cite{wang2024executable} uses executable code as the action space of LLM agents, and Thyme~\cite{zhang2025thymethinkimages} trains a multimodal model to write image-processing code while reasoning. To handle inputs far beyond the context window, Recursive Language Models (RLMs)~\cite{zhang2025recursive} store a long input as a variable in a persistent REPL and inspect it through code and recursive sub-calls, while MemGPT~\cite{packer2023memgpt} and context folding~\cite{sunscaling} manage long contexts with related paging and branching mechanisms. Text RLMs read the whole input inside the REPL at low cost. PMM builds on RLMs but reads days of video, where the program reads the temporal memory as text at low cost and pays a submodel call for each look at raw frames. Learning when to invoke costly tools has been explored in multimodal search~\cite{wu2025mmsearch} and model orchestration~\cite{su2025toolorchestra}. PMM instead learns when the temporal memory suffices and when to inspect raw frames, with reinforcement learning that uses answer correctness as the only reward, with no term for tool use or cost.

\section{Method}
\label{sec:method}

\begin{figure*}[t]
\centering
\includegraphics[width=\textwidth]{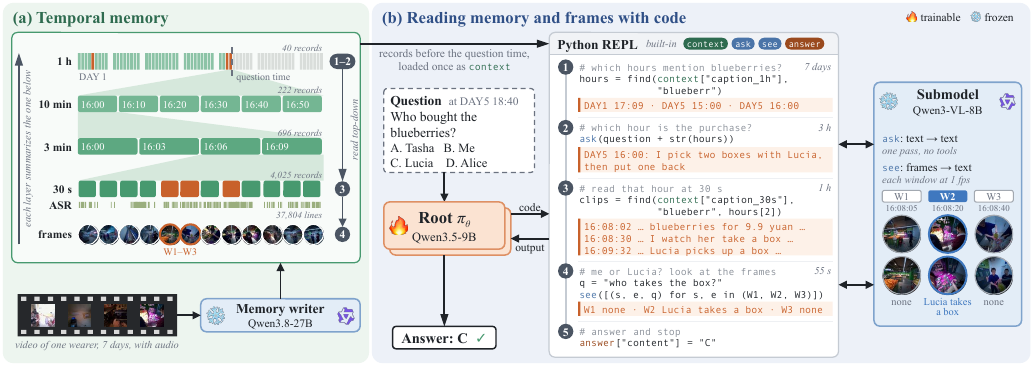}
\caption{\textbf{Overview of PMM.}
\textbf{(a)} \emph{Temporal memory.} A frozen VLM writer converts the video and transcript (ASR) into multi-scale captions that share one timeline with the raw frames. Records that end after the question time $t_{\mathrm{query}}$ (gray) are hidden.
\textbf{(b)} \emph{Reading memory and frames with code.} The visible memory is loaded into a persistent Python REPL as \texttt{context}. In each turn, the Root model $\pi_\theta$ writes code and reads only its output, filtering records locally, querying a frozen submodel on text (\texttt{ask}) or on frame windows (\texttt{see}), and setting \texttt{answer} to finish. Turn numbers in (a) mark what each turn reads, and orange denotes content reaching the Root. The trajectory in (b) is simplified, with \texttt{find} abbreviating a Python list comprehension over records.}
\label{fig:method}
\vspace{-10pt}
\end{figure*}

PMM couples a temporal memory of timestamped text with programmatic reading of this memory and of raw frames (\cref{fig:method}). After formalizing the problem (\cref{sec:method_setup}), we describe how PMM builds the multi-scale memory at write time without cross-event consolidation (\cref{sec:method_memory}), reads it and inspects frames with code in a persistent Python REPL (\cref{sec:method_access}), and trains an open-weight Root model directly in this REPL, so training matches inference (\cref{sec:method_training}).

\subsection{Problem Setup}
\label{sec:method_setup}

We consider multiple-choice question answering over a long video $V$, which may span days of continuous recording. Formally, $V$ is a stream of frames and aligned audio indexed on an absolute timeline $[0, T]$ in seconds. Given a question $q$ with candidate options $\mathcal{O}$ posed at question time $t_{\mathrm{query}} \in (0, T]$ ($t_{\mathrm{query}} = T$ when unspecified), the task is to output a prediction $\hat{y} \in \mathcal{O}$ for the ground-truth option $y^\ast \in \mathcal{O}$ using only the causal prefix $V_{\le t_{\mathrm{query}}}$. To avoid reprocessing $V$ from scratch for every question, we separate write time from read time. At write time, PMM converts $V$ once into a question-agnostic memory $\pmmMemory$. At read time, it answers each question from the causal portion of $\pmmMemory$ and from $V_{\le t_{\mathrm{query}}}$. The difficulty is that memory must be built before any question is known, while per-question reading cost must stay bounded as $V$ grows from hours to days.

\subsection{Temporal Memory}
\label{sec:method_memory}

At write time, PMM converts $V$ into a multi-scale temporal memory $\pmmMemory = \{\mathcal{L}_1, \ldots, \mathcal{L}_L, \mathcal{A}\}$ (\cref{fig:method}a), where $\mathcal{L}_1, \ldots, \mathcal{L}_L$ are $L$ caption layers at increasing temporal scales, $\mathcal{A}$ is the speech transcript, and each layer $S \in \pmmMemory$ stores time-ordered records $S = \{(t_j^0, t_j^1, x_j)\}_{j=1}^{|S|}$ with start time $t_j^0$, end time $t_j^1$, and text $x_j$ on the timeline of $V$. The transcript layer $\mathcal{A}$ uses official subtitles when available and automatic speech recognition otherwise. At the finest scale $\mathcal{L}_1$, $V$ is partitioned into non-overlapping 10\,s windows, and a frozen VLM writer receives the frames of the window together with overlapping lines from $\mathcal{A}$ to describe visible actions, people, objects, and speech. Each coarser layer (30\,s, 3\,min, 10\,min, and 1\,h, following the window scales of~\cite{yeo2026worldmm}) is generated from text alone, where each record summarizes the records of the layer below that start inside its window and spans their earliest start and latest end. Coarse layers let a program skim days of video in few records, and fine layers keep brief events that coarse summaries omit.

Because every layer summarizes only contiguous windows, $\pmmMemory$ stays aligned with the raw frames of $V$, allowing read-time programs to narrow down candidate intervals across scales and recover visual details missing from lossy captions. The memory contains no graph, embedding index, or entity table; any retrieval structure is computed by the program that reads it. At question time $t_{\mathrm{query}}$, the causal portion of $\pmmMemory$ is given by the visible memory $\pmmMemory_q = \{S_q \mid S \in \pmmMemory\}$, where $S_q = \{(t^0, t^1, x) \in S \mid t^1 \le t_{\mathrm{query}}\}$.

\subsection{Reading Memory and Frames with Code}
\label{sec:method_access}

At read time, PMM loads the visible memory $\pmmMemory_q$ once into a persistent Python REPL (\cref{fig:method}b) as a dictionary \texttt{context} mapping each layer name to its list of records, alongside a dictionary \texttt{answer}. Rather than placing $\pmmMemory_q$ in the prompt, we initialize the Root model $\pi_\theta$ with a compact prompt $h_1$ containing only the tool instructions, $q$, $\mathcal{O}$, and memory metadata, namely the layer names, record and character counts per layer, video duration, and $t_{\mathrm{query}}$ when specified. Any LLM that can write Python can serve as the Root without training; \cref{sec:method_training} trains an open-weight one. Reading then follows the multi-turn code execution of CodeAct~\cite{wang2024executable} and Recursive Language Models (RLMs)~\cite{zhang2025recursive}. At turn $r$, the Root generates a response $a_r$ from its history $h_r$, and the REPL executes the first Python code block $c_r$ in $a_r$:
\begin{equation}
\begin{aligned}
&a_r \sim \pi_\theta(\cdot \mid h_r), \qquad (z_{r+1}, o_r) = \mathrm{Exec}(c_r, z_r), \\
&h_{r+1} = h_r \oplus a_r \oplus \mathrm{Trunc}(o_r),
\end{aligned}
\label{eq:method_loop}
\end{equation}
where $z_r$ is the REPL state ($z_1$ holds the variables and functions in \cref{tab:method_tools}), $o_r$ is the printed output or Python exception message, $\mathrm{Trunc}$ retains a fixed-length prefix, and $\oplus$ denotes concatenation. Separating $z_r$ from $h_r$ decouples stored memory from neural context, since intermediate results such as candidate record lists stay in $z_r$ and enter $h_r$ only when printed. Exception messages also enter $h_{r+1}$, so the Root can repair faulty code in the next turn.

\begin{table}[t]
\centering\footnotesize
\setlength{\tabcolsep}{3pt}
\caption{\textbf{Variables and functions predefined in the Python REPL.} The Root combines them with arbitrary Python code, such as list comprehensions that filter records by time or keyword, sorting, counting, and loops over candidates. Examples follow \cref{fig:method}b. Both \texttt{ask} and \texttt{see} query the frozen submodel and take one request or a list of requests executed as one batched call.}
\label{tab:method_tools}
\begin{tabularx}{\columnwidth}{@{}ll>{\raggedright\arraybackslash}X@{}}
\toprule
Built-in & Example & Result\\
\midrule
\texttt{context} & \texttt{context["caption\_1h"]} & records of one layer\\
\texttt{ask} & \texttt{ask(question + str(hours))} & text\\
\texttt{see} & \texttt{see([(s, e, q), ...])} & text per window\\
\texttt{answer} & \texttt{answer["content"] = "C"} & ends the episode\\
\bottomrule
\end{tabularx}
\vspace{-10pt}
\end{table}

Unlike text RLMs, in which the full input resides in the REPL, \texttt{context} holds only a lossy temporal memory while $V_{\le t_{\mathrm{query}}}$ remains external on the same timeline. Root programs therefore combine Python code over \texttt{context} with two model calls of increasing cost, \texttt{ask} and \texttt{see} (\cref{tab:method_tools}); both use one frozen VLM as a stateless submodel that answers each query in one pass.

\noindent\textbf{Python code.}
Plain Python over \texttt{context}, such as filtering records by time or keyword, aligning layers, and counting events, runs locally in $z_r$ without any model call.

\noindent\textbf{Text queries.}
\texttt{ask} sends program-composed text, typically selected records paired with a sub-question, to the submodel and returns its text answer, offloading bulk record reading from the Root context.

\noindent\textbf{Frame queries.}
\texttt{see} takes one or a list of window queries $(t^0, t^1, q_{\mathrm{sub}})$. For each window, it samples a bounded number of uniformly spaced frames within $[t^0, t^1]$ from $V_{\le t_{\mathrm{query}}}$ and queries the submodel with $q_{\mathrm{sub}}$, a sub-question written by the Root. Answers return as text, so frame evidence enters the REPL in the same format as memory text.

Because these operations are ordinary Python functions, a single turn can filter thousands of records, loop over candidates, and issue multiple queries in one \texttt{ask} or \texttt{see} call (\cref{fig:method}b; \cref{fig:case_study} shows a full trajectory), whereas single-call tool interfaces require a separate turn per request. Rather than following a fixed pipeline, the Root decides for each question whether and where to inspect frames.

An episode $\tau = (a_1, o_1, \ldots, a_N, o_N)$ of $N$ turns ends when the program submits an option through \texttt{answer}, which becomes the prediction $\hat{y} \in \mathcal{O}$. Each episode is bounded by limits on turns, per-turn Root tokens, total Root context, and execution time (\cref{sec:experiments}). Reaching the turn limit without an answer triggers one extra turn that prompts the Root for $\hat{y}$ and is counted in $\tau$, whereas exhausting the token or time budget ends the episode without an answer, which counts as incorrect. The Root thus decides when to stop.

\subsection{Training the Root Model}
\label{sec:method_training}

We initialize the Root model $\pi_\theta$ from a pretrained open-weight LLM and train only $\theta$, keeping the memory, writer, and submodel frozen. Trajectory collection and RL run in the inference REPL with the same operations and budgets.

\noindent\textbf{Evidence-grounded SFT.}
A stronger teacher Root samples several trajectories per question on training videos disjoint from all evaluation benchmarks. We discard questions that the teacher answers correctly from $(q,\mathcal{O})$ alone, without memory or tools, since they do not test memory access. A trajectory is kept only if its answer is correct, it runs without tool or format errors, and it replays within the student's context budget. When the question annotates an evidence interval, the trajectory must also have observed it: a record overlapping the interval appeared in a printed output or in the input of an \texttt{ask} call, or a \texttt{see} call sampled frames inside it. This filter removes correct trajectories that never reached the evidence, so the student does not learn to answer without it. For questions whose correct option states that the answer cannot be determined, the teacher must choose that option consistently. We keep one trajectory per question and fine-tune $\pi_\theta$ with the next-token loss on all Root-generated tokens, including a forced final answer, while prompts and execution outputs are masked.

\noindent\textbf{Solvability-based curriculum.}
GRPO learns only from questions on which the policy sometimes succeeds. From a candidate pool $\mathcal{D}_{\mathrm{RL}}$, the SFT Root samples up to $G$ trajectories per question: questions with both correct and incorrect answers form $\mathcal{D}_{\mathrm{mix}}$, questions failed by all $G$ samples form $\mathcal{D}_{\mathrm{fail}}$, and questions always answered correctly are dropped. A first RL stage trains on $\mathcal{D}_{\mathrm{mix}}$. The resulting Root then samples $G$ trajectories for each question in $\mathcal{D}_{\mathrm{fail}}$; those that now yield mixed outcomes form the recovered set $\mathcal{D}_{\mathrm{rec}}$, on which a second stage continues from the first-stage weights. Questions in $\mathcal{D}_{\mathrm{fail}}$ give zero advantage to the SFT Root, so this step reuses those that the first stage makes learnable.

\noindent\textbf{Outcome-reward GRPO.}
In both stages, GRPO~\cite{shao2024deepseekmath} samples $G$ trajectories $\tau_1,\ldots,\tau_G$ per question and assigns each a binary reward and a group-normalized advantage:
\begin{equation}
  R_i=\mathbf{1}[\hat{y}(\tau_i)=y^\ast],
  \qquad
  \hat{A}_i=\frac{R_i-\bar{R}}{s_R+\varepsilon},
  \label{eq:method_reward}
\end{equation}
where $\hat{y}(\tau_i)$ is the submitted or forced answer of $\tau_i$, a missing answer or exhausted budget receives zero reward, and $\bar{R}$ and $s_R$ are the mean and standard deviation of the rewards in the group, and $\varepsilon>0$ is a small constant for numerical stability. Groups with equal rewards are kept with zero advantage. The clipped GRPO objective is applied to Root-generated tokens only. The reward has no term for format, tool use, or cost, so any preference between reading memory text and inspecting frames is learned only through its effect on answer correctness.

\section{Experiments}
\label{sec:experiments}
\subsection{Setup and Comparison Protocol}

\paragraph{Benchmarks and metrics.}
We evaluate personal memory on EgoLifeQA~\cite{yang2025egolife} (500 questions on week-long egocentric recordings of daily activities) and HippoVlog~\cite{lin2026hippomm} (960 questions across the 24 available videos), and long-video understanding on the MLVU test split~\cite{zhou2025mlvu} (502), LongVideoBench validation split~\cite{wu2024longvideobench} (1,337), and VideoMME Long~\cite{fu2025video} (900). We report accuracy on all benchmarks except MLVU, which uses its official nine-category macro-average (M-Avg), and summarize performance by the unweighted mean (Avg.)\ across the five benchmarks. Invalid or missing predictions are counted as incorrect, and published results are marked in \cref{tab:main}.

\paragraph{Implementation.}
The memory writer is Qwen3.8-27B~\cite{qwen2026qwen38}; on EgoLifeQA, HippoVlog, and VideoMME Long, some caption layers are instead imported from public releases. VideoMME Long is indexed without ASR transcripts. Reading uses Qwen3-VL-8B~\cite{bai2025qwen3vl} as the frozen submodel and greedy decoding for the Root. The writer reads frames at 1\,fps; each \texttt{see} window is sampled at 1\,fps and capped at 64 uniformly spaced frames. Each episode is capped at 20 turns, 4,096 new Root tokens per turn, and 65,536 total Root context tokens. Code execution times out after 300\,s and each question after 600\,s. Printed outputs are truncated at 20,000 characters. To train PMM-9B from Qwen3.5-9B, we fine-tune on 2,514 DeepSeek-V4.1-Flash teacher trajectories on SuperMemory-VQA~\cite{alam2026supermemory} for about 3 epochs. The RL pool has 3,000 questions from DramaVQA~\cite{huo2026dramavqa} and S-EMBER~\cite{wang2026sember}, plus a 200-question development set on separate videos. Screening and GRPO use $G=8$; screening yields 1,639 mixed-outcome and 578 all-failure questions, of which 108 are recovered after the first stage. We report the Root after the two RL stages (80 and 5 updates). Open-weight direct-frame baselines uniformly sample up to 128 frames preceding $t_{\mathrm{query}}$, while training-free memory and agent baselines are re-run with the backbones listed in \cref{tab:main} and trained agents with their released checkpoints, unless marked as published (full details in the supplement).

\subsection{Main Results}
\label{sec:main_results}

\begin{table*}[t]
\centering
\begingroup

\newcommand{\PMMAvg}[2]{%
    #1\textsuperscript{%
        \fontsize{6}{6}\selectfont
        \ifdim #2pt>0pt\relax
            \textcolor{green!50!black}{\ensuremath{#2}}%
        \else
            \textcolor{black!55}{\ensuremath{#2}}%
        \fi
    }%
}

\caption{
    \textbf{Main results on personal-memory and long-video benchmarks.}
    Accuracy (\%) on EgoLifeQA, HippoVlog, MLVU (test set; M-Avg over
    its nine task categories), LongVideoBench, and the long-video split
    of VideoMME.
    Avg.: mean over the five benchmarks.
    HippoVlog uses the 960 questions of the 24 videos available to us
    (the official release contains 25 videos and 1,000 questions).
    On VideoMME Long, PMM builds its memory without ASR transcripts.
    Our final model is shaded.
    The best and second-best results in each column are
    \textbf{bold} and \underline{underlined}, respectively.
    Superscripts give the difference from PMM-9B (zero-shot).
    $^\ast$Results reported in WorldMM~\cite{yeo2026worldmm},
    Molmo2~\cite{clark2026molmo2}, or the original method papers;
    all unmarked results are evaluated by us.
}
\label{tab:main}

\footnotesize
\setlength{\tabcolsep}{3pt}

\begin{tabularx}{\textwidth}{@{}Xll*{6}{c}@{}}
\toprule
& & & \multicolumn{2}{c}{Personal memory}
    & \multicolumn{3}{c}{Long-video understanding} & \\
\cmidrule(lr){4-5} \cmidrule(lr){6-8}
Method & Backbone & Source
    & EgoLifeQA & HippoVlog & MLVU & LongVideoBench & VideoMME Long
    & Avg. \\
& & & {\scriptsize $N{=}500$}
    & {\scriptsize $N{=}960$}
    & {\scriptsize $N{=}502$}
    & {\scriptsize $N{=}1{,}337$}
    & {\scriptsize $N{=}900$} & \\
\midrule

\multicolumn{9}{@{}l}{
    \textit{Direct-frame models (proprietary)}
} \\
GPT-5~\cite{openai2025gpt5}$^\ast$
    & -- & OpenAI
    & 48.6 & 75.7 & \underline{77.7} & 72.6 & \underline{74.3}
    & \PMMAvg{69.78}{+9.16} \\

Gemini 2.5 Pro~\cite{comanici2025gemini}$^\ast$
    & -- & Google
    & 46.4 & 72.0 & \textbf{81.5} & \textbf{76.8} & 55.7
    & \PMMAvg{66.48}{+5.86} \\
\midrule

\multicolumn{9}{@{}l}{
    \textit{Direct-frame models (open-source)}
} \\
Qwen3-VL-8B~\cite{bai2025qwen3vl}
    & 8B & Alibaba
    & 35.0 & 68.0 & 47.3 & 53.8 & 59.0
    & \PMMAvg{52.62}{-8.00} \\

Qwen3.5-9B~\cite{qwen2026qwen35}
    & 9B & Alibaba
    & 22.8 & 65.8 & 46.5 & 50.2 & 48.2
    & \PMMAvg{46.70}{-13.92} \\

Qwen2.5-VL-7B~\cite{bai2025qwen25vl}
    & 7B & Alibaba
    & 30.6 & 61.9 & 35.7 & 44.4 & 51.4
    & \PMMAvg{44.80}{-15.82} \\

InternVL3.5-8B~\cite{wang2025internvl35}
    & 8B & OpenGVLab
    & 32.0 & 72.1 & 47.6 & 57.7 & 53.1
    & \PMMAvg{52.50}{-8.12} \\
\midrule

\multicolumn{9}{@{}l}{
    \textit{Training-free memory and video agents}
} \\
WorldMM~\cite{yeo2026worldmm}
    & 8B & CVPR'26
    & 44.0 & 46.3 & 28.7 & 28.7 & 46.3
    & \PMMAvg{38.80}{-21.82} \\

HippoMM~\cite{lin2026hippomm}
    & 9B & CVPR'26
    & 32.4 & 75.7 & 42.3 & 53.0 & 44.7
    & \PMMAvg{49.62}{-11.00} \\

DVD~\cite{zhang2025dvd}
    & 9B & NeurIPS'25
    & 55.4 & 75.8 & 60.4 & 63.6 & 62.9
    & \PMMAvg{63.62}{+3.00} \\

VideoLucy~\cite{zuo2025videolucy}
    & 9B & NeurIPS'25
    & 53.4 & 75.8 & 47.4 & 59.6 & 63.2
    & \PMMAvg{59.88}{-0.74} \\

VideoTree~\cite{wang2025videotree}
    & 9B & CVPR'25
    & 39.6 & 73.6 & 51.9 & 53.5 & 61.8
    & \PMMAvg{56.08}{-4.54} \\
\midrule

\multicolumn{9}{@{}l}{
    \textit{Trained video-reasoning agents}
} \\
Ego-R1~\cite{tian2026ego}
    & -- & TPAMI'26
    & 53.0$^\ast$ & 58.8$^\ast$ & 43.0 & 34.5 & 64.9$^\ast$
    & \PMMAvg{50.84}{-9.78} \\

M3-Agent~\cite{long2026seeing}
    & -- & ICLR'26
    & 53.5$^\ast$ & 65.5$^\ast$ & 43.4 & 49.9 & 61.8$^\ast$
    & \PMMAvg{54.82}{-5.80} \\

LongVideo-R1~\cite{qiu2026longvideor1}
    & -- & CVPR'26
    & 34.2 & 67.7 & 56.0 & 58.0 & 57.3
    & \PMMAvg{54.64}{-5.98} \\

Video-MTR~\cite{xie2026videomtr}
    & -- & ICML'26
    & 31.4 & 59.8 & 47.8 & 57.1 & 51.3
    & \PMMAvg{49.48}{-11.14} \\

VideoExplorer~\cite{yuan2026videoexplorer}
    & -- & KDD'26
    & 30.6 & 63.1 & 49.5 & 56.1 & 42.7
    & \PMMAvg{48.40}{-12.22} \\
\midrule

\multicolumn{9}{@{}l}{
    \textit{PMM (ours)}
} \\
PMM-9B (zero-shot)
    & 9B & Ours
    & 49.6 & 74.9 & 54.5 & 65.8 & 58.3
    & \PMMAvg{60.62}{0.00} \\

PMM-9B (SFT)
    & 9B & Ours
    & 64.6 & 78.1 & 63.4 & 71.9 & 66.2
    & \PMMAvg{68.84}{+8.22} \\

\rowcolor{AblDefault}
\textbf{PMM-9B (SFT+RL)}
    & 9B & Ours
    & \underline{68.4}
    & \underline{80.8}
    & 68.6
    & 75.1
    & 70.6
    & \PMMAvg{\underline{72.70}}{+12.08} \\

\addlinespace[2pt]
PMM-DeepSeek
    & DeepSeek & Ours
    & \textbf{73.2}
    & \textbf{81.5}
    & 72.6
    & \underline{75.8}
    & \textbf{76.0}
    & \PMMAvg{\textbf{75.82}}{+15.20} \\
\bottomrule
\end{tabularx}

\endgroup
\vspace{-10pt}
\end{table*}

\begin{table}[t]
\centering
\caption{\textbf{Memory representation and placement.} \emph{w/o memory and Root}: Qwen3-VL-8B answers directly from 64 uniformly sampled frames. VideoMME Long has no ASR input, so \emph{w/o ASR} uses the default run on it.}
\label{tab:ablation}\label{tab:abl-memory}\label{tab:abl-placement}
\footnotesize
\setlength{\tabcolsep}{4pt}
\begin{tabularx}{\linewidth}{@{}Xrrr@{}}
\toprule
Configuration & Avg. & $\Delta$ & Tokens (K) \\
\midrule
\rowcolor{AblDefault}
Hierarchical memory + ASR, in REPL & \textbf{76.38} & -- & 118 \\
\midrule
\multicolumn{4}{@{}l}{\textit{Representation}} \\
w/o external memory       & 60.46 & $-15.92$ & 315 \\
w/o memory hierarchy      & 75.19 & $-1.18$  & 115 \\
Coarse-scale memory only  & 73.14 & $-3.23$  & 150 \\
w/o ASR transcripts       & \underline{75.51} & $-0.87$ & 119 \\
w/o memory and Root       & 49.40 & $-26.98$ & 11 \\
\midrule
\multicolumn{4}{@{}l}{\textit{Placement}} \\
In-prompt memory (32K)    & 71.49 & $-4.89$ & 315 \\
\bottomrule
\end{tabularx}
\vspace{-10pt}
\end{table}

Across the five benchmarks, PMM-9B (SFT+RL) achieves an average accuracy of 72.70\%, outperforming all compared baselines including standalone GPT-5 and Gemini 2.5 Pro (\cref{tab:main}). It leads GPT-5 and Gemini 2.5 Pro on the two personal-memory benchmarks, with margins of 5.1 and 8.8\,pp on HippoVlog and 19.8 and 22.0\,pp on week-long EgoLifeQA, where fixed frame budgets miss sparse, temporally distant evidence. On the three long-video benchmarks, PMM-9B remains competitive overall, whereas proprietary direct-frame models stay ahead on MLVU, whose shorter videos allow denser uniform frame coverage, as well as on LongVideoBench (Gemini 2.5 Pro) and VideoMME Long (GPT-5). Pairing PMM with the DeepSeek teacher Root requires no training and reaches the highest overall average at 75.82\%, ranking first on three of the five benchmarks.

Comparing against our open-weight backbones highlights the benefit of the memory and programmatic reading. Although PMM-9B uses Qwen3.5-9B as the Root and Qwen3-VL-8B as the submodel, its zero-shot variant already reaches 60.62\% on average, outperforming direct-frame inference with Qwen3.5-9B on every benchmark and Qwen3-VL-8B on four of the five. The largest gain is on EgoLifeQA (49.6\% vs.\ 22.8\% for Qwen3.5-9B), and the only loss is on VideoMME Long (58.3\% vs.\ 59.0\% for Qwen3-VL-8B), the only benchmark whose temporal memory has no speech transcripts. With the same Root and submodel, these gains come from building a memory of the full video and reading it selectively rather than from a stronger reader. Memory systems that consolidate the video at write time score lower: WorldMM, HippoMM, and M3-Agent average 38.80\%, 49.62\%, and 54.82\%, all below zero-shot PMM-9B, and reach at most 53.5\% on EgoLifeQA, where PMM-9B reaches 68.4\%.

Zero-shot PMM-9B also exceeds the average of every trained agent with complete results, suggesting that an unconsolidated temporal memory with programmatic reading is a strong starting point even before training. Still, multi-turn code search is demanding for an untrained 9B model, leaving zero-shot PMM-9B behind the strongest training-free agent (DVD) on four benchmarks. Two-stage Root training resolves this limitation, as SFT and RL each improve every benchmark. SFT adds 8.22\,pp on average, with the largest gain on week-long EgoLifeQA ($+15.0$\,pp), and RL adds 3.86\,pp, with the largest gain on MLVU ($+5.2$\,pp). The trained PMM-9B surpasses every training-free and trained agent on every evaluated benchmark, including DVD across all five sets, while narrowing the average gap to the teacher Root from 15.20 to 3.12\,pp. The Root is trained only on SuperMemory-VQA, DramaVQA, and S-EMBER, none of which we evaluate on.

\subsection{Ablation Studies and Analysis}
\label{sec:ablation}

\begin{figure}[t]
\centering
\includegraphics[width=0.82\linewidth]{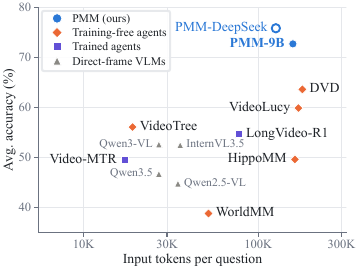}
\caption{
    \textbf{Accuracy and cost.}
    Average accuracy over the five benchmarks against input tokens per
    question over all model calls.
}
\label{fig:efficiency}
\vspace{-10pt}
\end{figure}

All ablations (\cref{tab:abl-memory,tab:abl-interface,tab:abl-budget,tab:abl-root}) use a 1,000-question subset with 200 questions each from EgoLifeQA, HippoVlog, MLVU (stratified by task category), LongVideoBench, and VideoMME Long (stratified by task type). Unless stated otherwise, we pair the DeepSeek-V4.1-Flash Root with the Qwen3-VL-8B submodel without training, and report five-benchmark average accuracy (M-Avg on MLVU) and total input tokens per question over Root turns, \texttt{see}, and \texttt{ask}. Shaded rows are the references for $\Delta$, the accuracy difference (pp), and the best and second-best Avg.\ in each table are in \textbf{bold} and \underline{underlined}.

\begin{table}[t]
\centering
\caption{\textbf{Action interface and tools.} \emph{JSON}: single-call tools with the same operations as code. \emph{Generic \texttt{see} queries}: every \texttt{see} sub-question is replaced by a fixed prompt to describe the frames.}
\label{tab:abl-interface}\label{tab:abl-tools}
\footnotesize
\setlength{\tabcolsep}{4pt}
\begin{tabularx}{\linewidth}{@{}l>{\raggedright\arraybackslash}Xrrr@{}}
\toprule
Root & Setting & Avg. & $\Delta$ & Tokens (K) \\
\midrule
\rowcolor{AblDefault}
DeepSeek-V4.1-Flash & Code            & \textbf{76.38} & -- & 118 \\
                    & JSON            & 75.02 & $-1.36$ & 121 \\
                    & w/o \texttt{see} & 73.89 & $-2.49$ & 87 \\
                    & w/o \texttt{ask} & 75.29 & $-1.09$ & 120 \\
                    & Generic \texttt{see} queries & \underline{76.01} & $-0.37$ & 122 \\
\midrule
\rowcolor{AblDefault}
Qwen3.8-27B         & Code            & 75.30 & -- & 111 \\
                    & JSON            & 73.00 & $-2.30$ & 105 \\
\bottomrule
\end{tabularx}
\vspace{-10pt}
\end{table}

\begin{table}[t]
\centering
\caption{\textbf{Search budget.} Adaptive search under different turn limits and a fixed 10-turn coarse-to-fine pipeline.}
\label{tab:abl-budget}
\footnotesize
\setlength{\tabcolsep}{4pt}
\begin{tabularx}{\linewidth}{@{}l*{4}{>{\raggedleft\arraybackslash}X}@{}}
\toprule
Strategy & Turns & Avg. & $\Delta$ & Tokens (K) \\
\midrule
Adaptive       & 2  & 60.81 & $-15.57$ & 11 \\
               & 5  & 68.30 & $-8.08$  & 32 \\
               & 10 & \underline{73.46} & $-2.91$ & 71 \\
\rowcolor{AblDefault}
               & 20 & \textbf{76.38} & --  & 118 \\
\midrule
Fixed pipeline & 10 & 67.58 & $-8.80$  & 95 \\
\bottomrule
\end{tabularx}
\vspace{-10pt}
\end{table}

\paragraph{Memory structure and placement.}
In \cref{tab:abl-memory}, the submodel alone with 64 uniform frames scores 49.40\%, adding the Root with \texttt{see} alone reaches 60.46\%, and full PMM reaches 76.38\%. External memory is the largest single factor ($+15.92$\,pp) and cuts total input tokens from 315K to 118K per question by localizing events in text before calling \texttt{see}. Retaining only coarse-scale captions lowers accuracy by 3.23\,pp and raises input tokens to 150K, whereas removing coarser summary layers or ASR transcripts causes smaller drops of 1.18 and 0.87\,pp. Where this memory resides also matters (\cref{tab:abl-placement}). Packing layers from coarse to fine into the initial prompt under a 32K-token budget (dropping any layer that does not fit as a whole while keeping \texttt{see} and \texttt{ask} available) lowers accuracy by 4.89\,pp and raises input tokens from 118K to 315K, mainly because the prompt is resent each turn. Keeping memory in the REPL preserves all scales while exposing only printed outputs to the Root context, regardless of memory size.

\paragraph{Programmatic interface and adaptive search.}
Programmatic reading benefits from both expressive actions and closed-loop control (\cref{tab:abl-interface,tab:abl-budget}). Executable code outperforms single-call JSON tools of matching capability by 1.36\,pp with DeepSeek-V4.1-Flash and 2.30\,pp with Qwen3.8-27B (\cref{tab:abl-interface}), as a single code block can compose filtering, aggregation, and temporal comparison while retaining intermediate state in the REPL. All four runs use 105K--121K input tokens per question, so code and JSON read similar amounts of input. Within the toolset (\cref{tab:abl-tools}), disabling \texttt{see} ($-2.49$\,pp) or \texttt{ask} ($-1.09$\,pp) lowers accuracy, consistent with the complementary roles of frame verification and submodel text offloading. Meanwhile, replacing every \texttt{see} sub-question with a fixed generic prompt to describe the frames barely changes accuracy ($-0.37$\,pp), indicating that visual gains come primarily from memory-guided window selection rather than question-specific prompting. Across turn budgets (\cref{tab:abl-budget}), accuracy rises monotonically from 60.81\% (2 turns) to 76.38\% (20 turns), while input tokens grow from 11K to 118K. With 2 turns, PMM uses 11K input tokens, the same as the submodel reading 64 uniform frames (\cref{tab:abl-memory}), yet is 11.41\,pp more accurate. We also compare against a 10-turn coarse-to-fine pipeline that restricts turns 1--3 to $\ge$3\,min layers, turns 4--6 to all text layers, and turns 7--9 to \texttt{see} before forcing a turn-10 answer, while allowing early answers throughout. Under the same 10-turn limit, adaptive search leads by 5.89\,pp with fewer input tokens (71K vs.\ 95K), as each turn builds on earlier outputs.

\begin{table}[t]
\centering
\caption{\textbf{Root and submodel.} $\Delta$ of a 27B-submodel row compares with the same Root using the 8B submodel.}
\label{tab:abl-root}
\footnotesize
\setlength{\tabcolsep}{4pt}
\begin{tabularx}{\linewidth}{@{}Xlrrr@{}}
\toprule
Root & Submodel & Avg. & $\Delta$ & Tokens (K) \\
\midrule
Qwen3.5-4B          & 8B  & 59.95 & $-16.43$ & 139 \\
Qwen3.5-9B          & 8B  & 62.12 & $-14.26$ & 125 \\
Qwen3-Coder-30B-A3B & 8B  & 59.69 & $-16.69$ & 173 \\
Qwen3.8-27B         & 8B  & 75.30 & $-1.08$  & 111 \\
\rowcolor{AblDefault}
DeepSeek-V4.1-Flash & 8B  & \underline{76.38} & -- & 118 \\
\midrule
DeepSeek-V4.1-Flash & 27B & \textbf{76.60} & $+0.22$ & 118 \\
Qwen3.5-9B          & 27B & 66.90 & $+4.78$ & 93 \\
\bottomrule
\end{tabularx}
\vspace{-10pt}
\end{table}

\begin{table}[t]
\centering
\caption{
    \textbf{Tool use of the Root before and after training}, per question,
    on the 1,000-question subset of \cref{tab:ablation}, averaged over the five benchmarks.
    Root in/out: input/output tokens of Root turns (K), excluding \texttt{see} and \texttt{ask};
    Fr.: frames per question; Forced: episodes ending with a forced answer or exhausted budget (\%).
}
\label{tab:training_behavior}
\footnotesize
\setlength{\tabcolsep}{3pt}
\resizebox{\columnwidth}{!}{%
\begin{tabular}{@{}lrrrrrrrr@{}}
\toprule
Root & Avg. & Turns & \texttt{see} & Fr. & \texttt{ask} & Root in & Root out & Forced \\
\midrule
PMM-9B (zero-shot) & 62.12 & 10.0 & 4.01 & 48 & 1.19 & 104 & 4.08 & 4.10 \\
PMM-9B (SFT)      & 69.94 & 9.2 & 6.60 & 135 & 0.19 & 99 & 2.06 & 0.30 \\
PMM-9B (SFT+RL)   & 73.31 & 10.3 & 7.62 & 170 & 0.23 & 117 & 2.84 & 0.00 \\
\midrule
PMM-DeepSeek & 76.38 & 8.4 & 9.26 & 150 & 0.32 & 87 & 1.96 & 0.30 \\
\bottomrule
\end{tabular}%
}
\vspace{-10pt}
\end{table}

\begin{figure*}[t]
\centering
\includegraphics[width=0.9\textwidth]{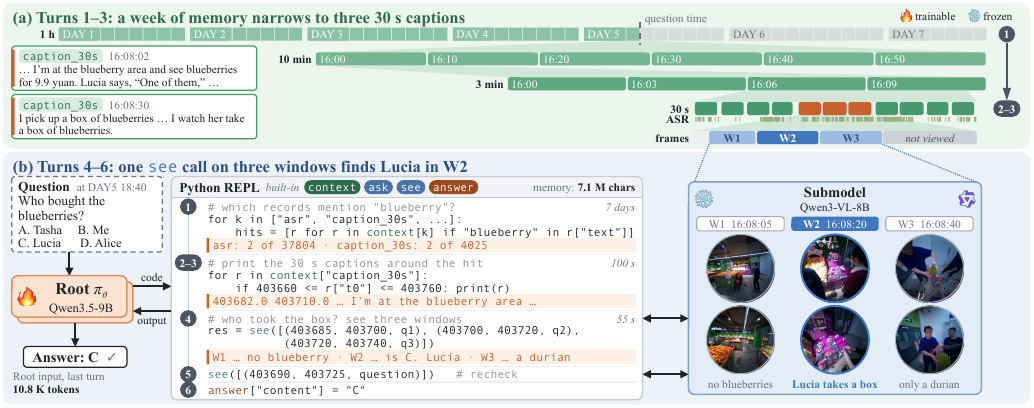}
\caption{\textbf{A real trajectory of PMM-9B on EgoLifeQA.} Orange marks the 30\,s records printed in turns 2--3; the magnified frames row shows the three \texttt{see} windows, and only W2 shows the purchase. Code and outputs are abridged, \texttt{\#} comments summarize the code, and clock times are converted from seconds.}
\label{fig:case_study}
\vspace{-10pt}
\end{figure*}

\paragraph{Root vs.\ submodel capacity.}
Varying the two models independently shows that reasoning, rather than perception, is the primary bottleneck (\cref{tab:abl-root}). With the 8B submodel fixed, upgrading the Root from Qwen3.5-9B to Qwen3.8-27B boosts accuracy by 13.18\,pp, whereas Qwen3-Coder-30B-A3B~\cite{yang2025qwen3}, an MoE coding model with about 3B active parameters, scores below the dense 9B Root (59.69\% vs.\ 62.12\%), indicating that neither total parameter count nor coding specialization substitutes for general reasoning. Scaling the submodel from 8B to 27B adds only 0.22\,pp under DeepSeek-V4.1-Flash, while Qwen3.5-9B gains 4.78\,pp with the larger submodel. Planning and coordinating evidence thus matter more than submodel size, which supports training only the Root.

\paragraph{Efficiency.}\label{sec:efficiency}
\Cref{fig:efficiency} compares accuracy (main-table runs) with input tokens per question on all five benchmarks.
PMM-9B (SFT+RL) is more accurate than every baseline, at 72.70\% with 158K tokens, and it uses fewer tokens than DVD (63.62\%, 179K), VideoLucy (59.88\%, 170K), and HippoMM (49.62\%, 162K); with the teacher Root, PMM reaches 75.82\% at 126K. Turn limits also trade accuracy for cost: with 5 turns, the teacher Root uses 32K input tokens, comparable to direct-frame VLMs, and reaches 68.30\% on the ablation subset (\cref{tab:abl-budget}).
The remaining methods are cheaper but at least 16.6\,pp less accurate: LongVideo-R1 (54.64\%, 77K), WorldMM (38.80\%, 52K), VideoTree (56.08\%, 19K), Video-MTR (49.48\%, 17K), and the open-source direct-frame VLMs (at most 52.62\%, 27K--36K).

\paragraph{Effect of training.}\label{sec:training_analysis}
\Cref{tab:training_behavior} shows that training shifts the 9B Root from \texttt{ask} to \texttt{see}.
The zero-shot Root calls \texttt{ask} 1.19 times per question, inspects 48 frames, and ends 4.10\% of questions forced or over budget; after SFT, \texttt{see} calls rise from 4.01 to 6.60 and frames from 48 to 135, while \texttt{ask} calls fall to 0.19 and forced endings to 0.30\%. Root output also halves from 4.08K to 2.06K tokens per question. Turns stay between 9.2 and 10.3 at every stage, so training changes what each turn does, not how long the Root searches.
RL further raises \texttt{see} calls and frames (7.62 calls, 170 frames), ends every episode with an answer, and raises accuracy on the 1,000-question subset from 69.94\% to 73.31\%, against 76.38\% for the teacher Root. Training thus moves the 9B Root toward the teacher, which makes 9.26 \texttt{see} and 0.32 \texttt{ask} calls per question.

\paragraph{Case study.}
\Cref{fig:case_study} shows a real EgoLifeQA trajectory of PMM-9B. In turns 1--3, the Root searches every layer for ``blueberry'' and prints the captions and transcript around the hit. The text narrows a week of video to about one minute but does not tell who bought the blueberries. In turn 4, one program sends three windows to the submodel in a single \texttt{see} call, and only W2 shows Lucia taking a box. A second \texttt{see} call over this moment confirms Lucia in turn 5, and the Root answers correctly in turn 6. Text localizes the event, and frames resolve what captions omit. The 7.1M-character temporal memory stays in the REPL, while the Root input at the last turn is 10.8K tokens.

\section{Conclusion}
We presented PMM, which decides what matters in long videos at read time rather than at write time. PMM keeps an unconsolidated temporal memory aligned with the raw video, and a trained Root model reads it with code, inspecting raw frames only where the text points. With a 9B Root, PMM outperforms every open-source model and agent baseline on five benchmarks and beats standalone GPT-5 and Gemini 2.5 Pro on average. These results suggest that a long-horizon memory need not anticipate future questions when a trained reader can search it.

{
    \small
    \bibliographystyle{ieeenat_fullname}
    \bibliography{main}
}

\end{document}